%% file: main.tex
\documentclass[10pt,twocolumn,letterpaper]{article}

\usepackage{cvpr}

\usepackage{multirow}
\usepackage{graphicx}

\definecolor{cvprblue}{rgb}{0.21,0.49,0.74}
\usepackage[pagebackref,breaklinks,colorlinks,allcolors=cvprblue]{hyperref}

\def\paperID{*****}
\def\confName{CVPR}
\def\confYear{2025}

\title{Zero-Shot Gaze-based Volumetric Medical Image Segmentation}

\author{
Tatyana Shmykova \quad Leila Khaertdinova \quad Ilya Pershin \\
Research Center of the Artificial Intelligence Institute, Innopolis University, Innopolis, Russia \\
{\tt\small \{t.shmykova,i.pershin\}@innopolis.ru, l.khaertdinova@innopolis.university}
}

\begin{document}
\maketitle

\input{sections/0_abstract}    
\input{sections/1_intro}
\input{sections/2_related}
\input{sections/3_methods}
\input{sections/4_results}
\input{sections/5_conclusion}
{
    \small
    \bibliographystyle{unsrt}
    \bibliography{main}
}

\end{document}

%% file: sections/0_abstract.tex
\begin{abstract}

Accurate segmentation of anatomical structures in volumetric medical images is crucial for clinical applications, including disease monitoring and cancer treatment planning. Contemporary interactive segmentation models, such as Segment Anything Model 2 (SAM-2) and its medical variant (MedSAM-2), rely on manually provided prompts like bounding boxes and mouse clicks. In this study, we introduce eye gaze as a novel informational modality for interactive segmentation, marking the application of eye-tracking for 3D medical image segmentation. We evaluate the performance of using gaze-based prompts with SAM-2 and MedSAM-2 using both synthetic and real gaze data. Compared to bounding boxes, gaze-based prompts offer a time-efficient interaction approach with slightly lower segmentation quality. Our findings highlight the potential of using gaze as a complementary input modality for interactive 3D medical image segmentation.
\end{abstract}

%% file: sections/1_intro.tex
\section{Introduction}

Accurate anatomical segmentation is vital for cancer diagnosis, treatment planning, and disease monitoring. Although automated segmentation methods have shown promise, they often struggle with the variability of medical images, necessitating interactive approaches that incorporate user input for enhanced accuracy and adaptability. Recent advancements in interactive segmentation have demonstrated the potential to combine automated algorithms with different forms of user guidance. For instance, Segment Anything Model (SAM) \cite{kirillov2023segment} and its medical variant, MedSAM \cite{medsam}, leverage user prompts such as bounding boxes, clicks, and masks to refine segmentation results, enhancing both precision and efficiency.

Various prompting techniques have been explored for the medical image segmentation task, including bounding boxes (MedSAM \cite{medsam} and MedSAM-2 \cite{medsam2}), text prompts (MedCLIP-SAMv2 \cite{medclip}), and gaze-based inputs (GazeSAM \cite{gazesam} and \cite{gazeassisted,ieeegaze}). Existing approaches often rely on fine-tuning to adapt to new prompting modalities, such as gaze inputs. However, these methods require additional labeled data, which may be scarce for narrow-domain tasks.
Foundation models, such as MedSAM \cite{medsam} and MedSAM-2 \cite{medsam2}, take a medical image as input along with a specifies region of interest. Unlikely, multimodal models, including MedCLIP-SAMv2 \cite{medclip} integrate information from multiple modalities (images and text), allowing them to interpret visual data within the context of textual descriptions.
Despite the MedCLIP-SAMv2 claimed performance in zero-shot settings \cite{medclip}, we observed that text prompts perform poorly when segmenting abdominal organs in the WORD dataset \cite{luo2022word}. This highlights the need for an alternative zero-shot approach that enables MedSAM to be used effectively on specific medical segmentation tasks without additional fine-tuning. Our proposed method addresses this limitation by providing an efficient and accurate segmentation solution tailored to narrow-domain applications.

One of the major challenges in medical imaging is the transition from segmentation of 2D medical images to 3D images, as annotation of volumetric scans performed per slice can be extremely time-consuming. Recent models, such as SAM-2 \cite{sam2} and MedSAM-2 \cite{medsam2}, have introduced memory mechanisms to facilitate interactive 3D segmentation by incorporating prior contextual information. However, traditional input methods like manual clicks and bounding boxes remain discrete and require manual effort, which can be inefficient for large-scale volumetric data.

\begin{figure*}[t]
  \centering
   \includegraphics[width=\linewidth]{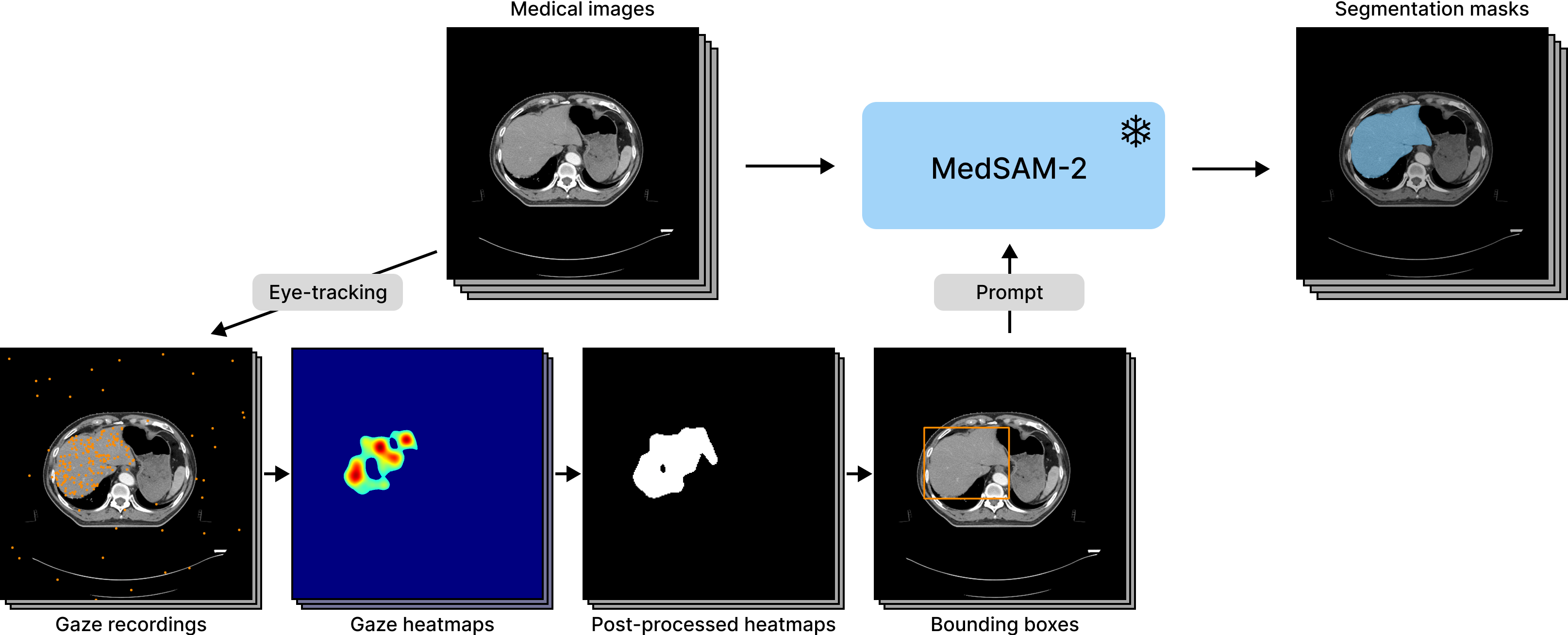}
   \caption{The process of generating segmentation masks based on the annotator's gaze stream involves the following steps: (1) tracking of eye gaze, (2) gaze heatmaps generation, (3) heatmaps post-processing, (4) bounding box-based prompt construction, and (5) segmentation using a pre-trained interactive segmentation model, e.g., MedSAM-2, with frozen parameters.}
   \label{method}
\end{figure*}

In this study, we propose gaze as a novel informational modality for interactive 3D medical image segmentation. Unlike conventional user prompts, gaze is a continuous and passive input that does not require explicit manual intervention, making it a more natural and intuitive interaction method. While gaze has been previously explored as a prompting technique for segmentation of 2D images, its potential for 3D segmentation remains largely unexplored. We bridge this gap by integrating gaze-based inputs into interactive segmentation models and comparing their performance and efficiency with traditional prompting techniques.

To assess the feasibility and effectiveness of our approach, we conduct experiments using a workstation equipped with a Tobii Eye Tracker 4C, allowing an expert radiologist to interactively segment medical images using gaze-based prompts. Our results demonstrate that gaze-based segmentation offers a faster and more seamless alternative to bounding boxes-based method with minimal loss in accuracy. This study presents eye gaze as an innovative modality for interactive 3D medical image segmentation, paving the way for more intuitive and efficient approach for human-AI collaboration in the medicine domain.  

%% file: sections/2_related.tex
\section{Related work}

The zero-shot segmentation method in MedCLIP-SAMv2 integrates multimodal learning to achieve precise medical image segmentation without requiring manual annotations. The approach begins with BiomedCLIP, a specialized vision-language model, which extracts both image and text embeddings. These embeddings are refined through the Multi-modal Information Bottleneck (M2IB) mechanism, which enhances relevant visual information while suppressing noise. The resulting saliency maps undergo a postprocessing step with K-Means clustering to generate an initial segmentation mask. This mask serves as the basis for creating bounding box or point-based prompts for the Segment Anything Model (SAM), which further refines the segmentation output.  

The authors conduct experiments to assess the impact of text prompt design on zero-shot segmentation. They compare six different types of text prompts: P0 and P1 use class names, P2 and P3 feature descriptive prompts, and P4 and P5 employ ensembles of 20 prompts. P0, P2, and P4 are generic, whereas P1, P3, and P5 specify object subtypes. For example, in Breast Ultrasound, P0 is "breast tumor," while P1 specifies "malignant" or "benign" tumors. P2 provides a single descriptive sentence, whereas P3 details tumor subtypes. P4 and P5 follow the same structure but use ensemble embeddings. GPT-4 generates all descriptive prompts. The authors evaluate their approach on various datasets, including Lung CT, where the lowest Dice score is 69.89 when using P0 prompts, which contain only the class name. With P2 prompts, the performance improves to 80.38.

To assess the performance of the zero-shot segmentation algorithm MedCLIP-SAMv2, we use two types of text prompts: P0, which use the class name of the organ, and P2, which consists of descriptive sentence prompts in the format: "A medical CT scan showing the \textit{class name of the organ}. Although the authors of MedCLIP-SAMv2 claim that their method demonstrates strong zero-shot capabilities, its performance on the WORD dataset is notably poor. Our evaluation is conducted on 600 slices of the WORD dataset, employing MedCLIP-SAMv2 approach. While the model achieves a Dice score of 0.88 when using bounding boxes derived from ground truth mask boundaries, its performance drops significantly when using P0 and P2 text prompts, yielding scores of 0.16 and 0.2, respectively. These results indicate that the proposed approach is not effective across all datasets.

%% file: sections/3_methods.tex
\section{Methods}

3D interactive medical image segmentation focuses on accurately delineating organs or pathological structures across all slices of a volumetric scan. The input is a 3D medical image represented as a stack of slices, and the output consists of segmentation masks for each slice. Unlike fully automated segmentation, interactive segmentation incorporates user input, such as bounding boxes, clicks, or other prompts to refine and guide the process. These prompts may be applied to all slices or only a subset. The challenge lies in ensuring accurate, consistent, and efficient segmentation with minimal user interaction while adapting to variations in anatomy and pathology.

Our method consists of the following steps: gaze tracking using an eye tracker, generation of gaze heatmaps, postprocessing of heatmaps, and generation of bounding boxes to obtain the final segmentation mask using a pretrained model, as illustrated in Figure~\ref{method}.

\textbf{Gaze heatmaps generation.}
To further process the gaze data, collected by eye tracking, we constructed a 2D gaze heatmap using an algorithm based on a Gaussian filter. We employed the same post-processing method as MedCLIP-SAMv2 \cite{medclip}, replacing text saliency maps with our generated gaze heatmaps. This method includes K-Means clustering to produce a coarse segmentation mask.

\textbf{Segmentation Refinement via SAM-2.}
The initial segmentation is passed as an input into SAM-2, which refines the segmentation using visual prompts derived from the coarse segmentation mask. In our approach, we use gaze-based bounding boxes as prompts for the SAM-2 model, calculating boxes to enclose each connected contour in the initial segmentation. For all selected CT slices, bounding boxes are provided as prompts to the SAM-2 model.

\textbf{Shape-based interpolation between boolean masks.}
To generate all masks throughout the CT scan we used interpolation between boolean masks, described in article \cite{schenk2000efficient}. The shape-based approach creates a binary scene from an object contour, applies a distance transformation, and interpolates the transformed images using grayscale techniques. Finally, zero-crossings then convert them back into binary contours. The distance transformation is computed using two successive chamfering processes with 3×3 kernel operations, approximating the Euclidean distance.

%% file: sections/4_results.tex
\section{Experiments and Results}

The segmentation performance of SAM-2 and MedSAM-2 models is evaluated on both synthetic and real gaze data, using different prompting strategies. Specifically, we compare gaze prompts with bounding boxes (bboxes) synthetically generated based on ground truth mask boundaries, as well as real bounding boxes provided by the medical expert. We also estimate the average time needed for the expert to segment an organ on a single 3D CT scan.

To validate our approach on real-world data, we invited a proxy radiologist. Before evaluation, the proxy radiologist underwent training in abdominal organ segmentation under the supervision of an experienced radiologist. This training ensured familiarity with anatomical structures and segmentation guidelines, allowing for a more accurate and consistent evaluation of our method.

\subsection{Experimental Setting}

\textbf{Gaze tracking.}
We develop a radiologist workstation with integrated eye-tracking functionality, designed for use in a dedicated, isolated room at our institution. The workstation is equipped with a lightweight, user-friendly, bar-shaped eye-tracking device positioned beneath the monitor for convenience. The hardware setup includes an LG diagnostic monitor featuring 10-bit color depth, a resolution of 3840×2160 pixels, and a pixel density of 7.21 px/mm. The eye-tracking functionality is facilitated by a Tobii Eye Tracker 4C, which operates at a frequency of 90 Hz.

\begin{figure}[t]
  \centering
   \includegraphics[width=\linewidth]{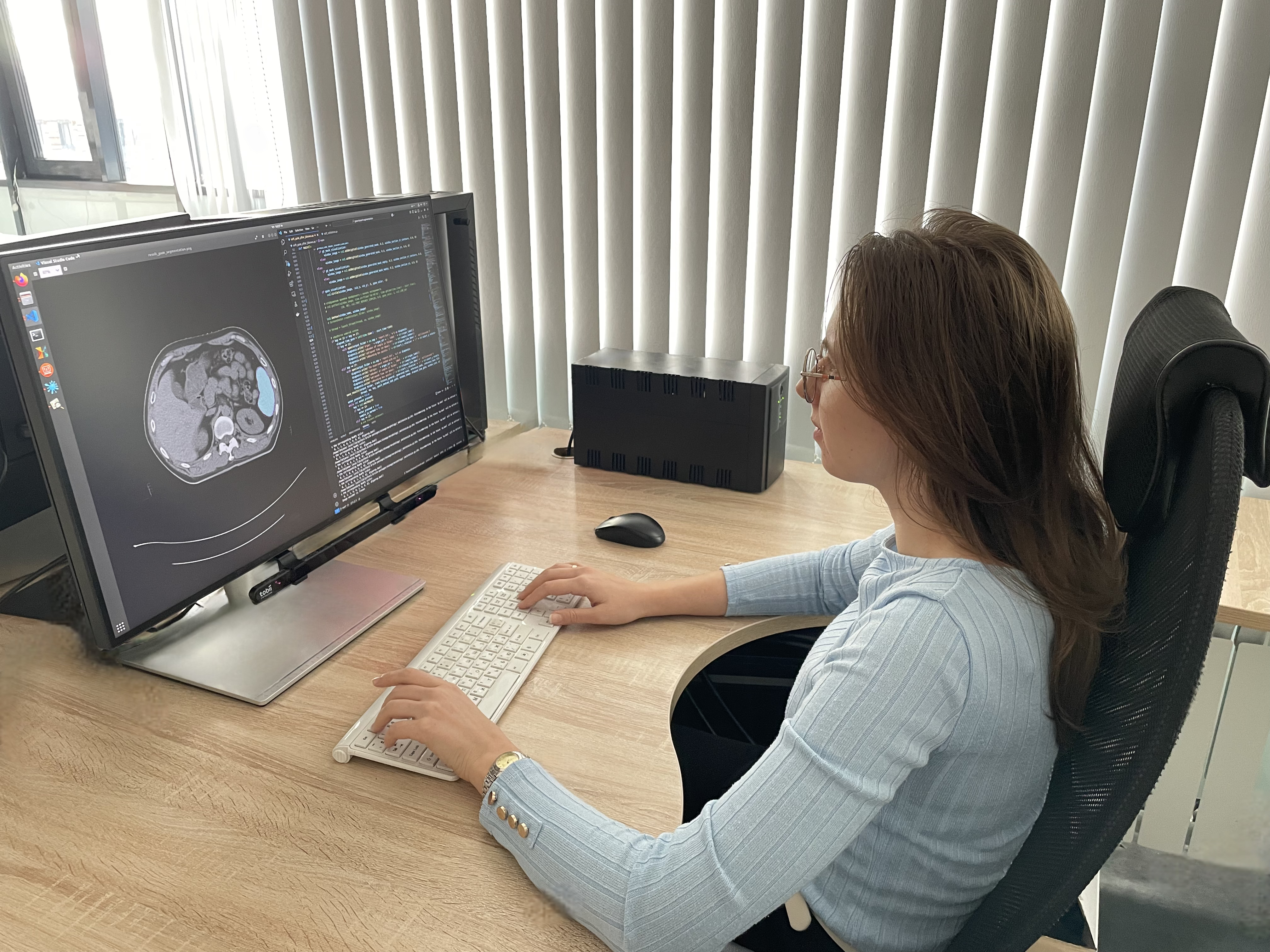}

   \caption{Gaze-based system that enables radiologists to segment abdominal organs on CT scans using an eye-tracker.}
   \label{proxy}
\end{figure}

\textbf{Dataset.}
We utilize the WORD dataset \cite{luo2022word}, encompassing abdominal volumetric CT images, to explore using the annotator's gaze as prompt for SAM-2 and MedSAM-2. This dataset comprises 150 CT scans from 150 patients, covering 16 abdominal organs. Each CT scan contains between 159 and 330 slices, each with a resolution of 512 × 512 pixels. For experiments, we use 350 CT slices of 16 abdominal organs.

\textbf{Synthetic gaze data.}
For testing models on synthetic data, we generate gaze points by sampling coordinates from reference segmentation masks for abdominal organs in 2D CT slices from WORD \cite{luo2022word}. Specifically, 80\% of the points are randomly generated inside each organ area and 20\% outside the organ, simulating natural gaze fluctuations and assuming potential inaccuracies in eye-tracking data \cite{ieeegaze}.

\subsection{Different strategies for prompt}

\begin{table}[ht]
\centering
\begin{tabular}{@{}lccc@{}}
\toprule
Model                     & Method      & Dice score           & Time (sec)    \\
\midrule
\multirow{3}{*}{SAM-2}    & First slice                 & 0.661 $\pm$ 0.318 & 12 $\pm$ 1  \\
                          & All slices & 0.896 $\pm$ 0.073 & 133 $\pm$ 4 \\
                          & 30 slices  & \bf 0.896 $\pm$ 0.073 & \bf 102 $\pm$ 4 \\
\hline
\multirow{3}{*}{MedSAM-2} & First slice                 & 0.670 $\pm$ 0.353 & 11 $\pm$ 1  \\
                          & All slices & 0.904 $\pm$ 0.063 & 133 $\pm$ 4 \\
                          & 30 slices  & \bf 0.904 $\pm$ 0.063 & \bf 102 $\pm$ 4 \\
\bottomrule
\end{tabular}
\caption{Comparison of using different numbers of slices on which the prompt is provided on synthetic data: (1) only on the first slice, (2) on all slices, and (3) on 30 slices. Prompts generated based on \textbf{ground truth mask} boundaries.}
\label{synthetic-bbox}
\end{table}

\begin{table}[ht]
\centering
\begin{tabular}{@{}lccc@{}}
\toprule
Model                     & Method  & Dice score           & Time (sec)    \\
\midrule
\multirow{2}{*}{SAM-2}    &   All slices            & 0.809 $\pm$ 0.171 & 108 $\pm$ 7 \\
                          &   30 slices             & \bf 0.812 $\pm$ 0.172 & \bf 82 $\pm$ 7  \\
\hline
\multirow{2}{*}{MedSAM-2} &   All slices            & 0.814 $\pm$ 0.163 & 107 $\pm$ 7 \\
                          &   30 slices             & \bf 0.817 $\pm$ 0.162 & \bf 82 $\pm$ 7 \\
\bottomrule
\end{tabular}
\caption{Comparison of using different numbers of slices on which the prompt is provided on synthetic data: (1) only on the first slice, (2) on all slices, and (3) on 30 slices. Prompts generated based on \textbf{synthetic gaze heatmaps}.}
\label{synthetic-gaze}
\end{table}

We test different synthetic prompt strategies, changing the number of slices on which prompts are provided. According to Tables \ref{synthetic-bbox} and \ref{synthetic-gaze}, when limited to the first slice, segmentation performance decreases substantially for both models (0.661 for SAM-2 and 0.670 for MedSAM-2). Limiting the number of prompts to 30 slices maintains similar accuracy compared to providing prompts on all images (0.896 for SAM-2 and 0.904 for MedSAM-2 using bbox) and significantly reduces segmentation time (133 vs. 102 seconds using bbox).

\subsection{Effectiveness of Gaze prompts}

\begin{table}[ht]
\centering
\begin{tabular}{@{}lcccc@{}}
\toprule
Model                     & Prompts & Dice                         & Time (sec)               \\
\midrule
\multirow{2}{*}{SAM-2}    & Bbox    & \bf 0.834 $\pm$ 0.124 & 88 $\pm$ 4              \\
                          & Gaze    & 0.750 $\pm$ 0.204              & \bf 63 $\pm$ 7 \\
\hline
\multirow{2}{*}{MedSAM-2} & Bbox    & \bf 0.844 $\pm$ 0.115 & 88 $\pm$ 4              \\
                          & Gaze    & 0.759 $\pm$ 0.203              & \bf 62 $\pm$ 7 \\
\bottomrule
\end{tabular}
\caption{Comparison of segmentation performance and mean time between SAM-2 and MedSAM-2 on real prompts provided by the proxy radiologist. Prompt methods: Bbox (bounding boxes), Gaze (prompts generated based on gaze heatmaps).}
\label{real-sam}
\end{table}

According to Table \ref{real-sam}, both models achieve higher Dice scores when using synthetic bounding boxes, with MedSAM-2 slightly outperforming SAM-2 (0.904 vs. 0.896). Synthetic gaze-based prompts, while slightly less accurate than bounding boxes, significantly reduce the time to get the masklet (82 vs. 102 secs). Similarly, bounding boxes drawn by the radiologist results in better segmentation performance (0.834 for SAM-2, 0.844 for MedSAM-2) compared to gaze-based prompts (0.750 and 0.759, respectively). However, using gaze-based prompts remains efficient, requiring less time than manual bounding boxes (62 vs. 88 secs). 

%% file: sections/5_conclusion.tex
\section{Conclusion and Future work}

In this work, we introduce and evaluate a gaze-based approach for 3D interactive medical image segmentation, which does not rely on fine-tuned interactive segmentation models. Gaze-based prompts represent a novel form of multimodal interaction. Unlike discrete prompts like clicks, gaze provides continuous input, making interaction more natural. Our approach builds upon previous research that utilized gaze for 2D segmentation by adapting this concept for 3D images. While bounding box-based methods (SAM-2, MedSAM-2) achieve the highest accuracy, gaze-based prompting offer a faster alternative with minimal accuracy loss. MedSAM-2 outperform SAM-2 on synthetic and real data, highlighting its robustness.

Despite the findings, the experiments with a proxy radiologist may not fully reflect real-world clinical conditions. Future work should validate results with clinicians and explore the generalizability of gaze-based prompting across diverse medical imaging datasets, incorporating different anatomies and imaging modalities.

\section*{Acknowledgement}

All authors were supported by the Research Center of the Artificial Intelligence Institute at Innopolis University. Financial support was provided by the Ministry of Economic Development of the Russian Federation (No. 25-139-66879-1-0003).